\documentclass[lettersize,journal]{IEEEtran}
\usepackage{amsmath,amsfonts}
\usepackage{algorithmic}
\usepackage{algorithm}
\usepackage{array}
\usepackage[caption=false,font=normalsize,labelfont=sf,textfont=sf]{subfig}
\usepackage{textcomp}
\usepackage{stfloats}
\usepackage{url}
\usepackage{verbatim}
\usepackage{graphicx}
\usepackage{orcidlink}
\usepackage{xcolor}

\usepackage[style=ieee]{biblatex}  
\usepackage[utf8]{inputenc}

\usepackage{booktabs}

\usepackage{multirow}

\begin{document}

\title{SynA-ResNet: Spike-driven ResNet Achieved through OR Residual Connection}

\author{Yimeng~Shan\orcidlink{0009-0000-1952-0119}, ~\IEEEmembership{Graduate~Student~Member,~IEEE}, Xuerui Qiu\orcidlink{0009-0008-3734-4671}, ~\IEEEmembership{Graduate~Student~Member,~IEEE}, Rui-Jie Zhu, ~\IEEEmembership{Graduate~Student~Member,~IEEE}, Jason K. Eshraghian\orcidlink{0000-0002-5832-4054}, ~\IEEEmembership{Member,~IEEE}, Malu Zhang\orcidlink{0000-0002-2345-0974}, ~\IEEEmembership{Member,~IEEE},  Haicheng Qu\orcidlink{0000-0002-2351-3154}, ~\IEEEmembership{Member,~IEEE}

\thanks{This work was partly supported by the National Natural Science Foundation of China under Grant 42271409, and the Scientific Research Fund of Liaoning Provincial Education Department under Grant LIKMZ20220699. (\textit{Corresponding author: Haicheng Qu})}
\thanks{Yimeng Shan and Haicheng Qu are with the School of Software, Liaoning Technical University, Huludao, China (e-mail: yimengshan2001@gmail.com; quhaicheng@lntu.edu.cn).}
\thanks{Xuerui Qiu is with the Institute of Automation, Chinese Academy of Sciences, Beijing, China (e-mail: qiuxuerui2024@ia.ac.cn).}
\thanks{Rui-Jie Zhu and Jason K. Eshraghian is with the University of California, Santa Cruz, Santa Cruz, CA USA (e-mail: rzhu48@ucsc.edu; jeshraghian@gmail.com).}
\thanks{Malu Zhang is with the University of Electronic Science and Technology of China, Chengdu, China (e-mail: maluzhang@uestc.edu.cn).}

}


\markboth{Journal of \LaTeX\ Class Files,~Vol.~14, No.~8, August~2021}%
{Shell \MakeLowercase{\textit{et al.}}: A Sample Article Using IEEEtran.cls for IEEE Journals}

\IEEEpubid{This work has been submitted to the IEEE for possible publication. Copyright may be transferred without notice, after which this version may no longer be accessible.}


\maketitle

\begin{abstract}

Spiking Neural Networks (SNNs) have garnered substantial attention in brain-like computing for their biological fidelity and the capacity to execute energy-efficient spike-driven operations. As the demand for heightened performance in SNNs surges, the trend towards training deeper networks becomes imperative, while residual learning stands as a pivotal method for training deep neural networks. In our investigation, we identified that the SEW-ResNet, a prominent representative of deep residual spiking neural networks, incorporates non-event-driven operations. To rectify this, we propose a novel training paradigm that first accumulates a large amount of redundant information through OR Residual Connection (ORRC), and then filters out the redundant information using the Synergistic Attention (SynA) module, which promotes feature extraction in the backbone while suppressing the influence of noise and useless features in the shortcuts. When integrating SynA into the network, we observed the phenomenon of "natural pruning", where after training, some or all of the shortcuts in the network naturally drop out without affecting the model's classification accuracy. This significantly reduces computational overhead and makes it more suitable for deployment on edge devices. Experimental results on various public datasets confirmed that the SynA-ResNet achieved single-sample classification with as little as 0.8 spikes per neuron. Moreover, when compared to other residual SNN models, it exhibited higher accuracy and up to a 28-fold reduction in energy consumption. Codes are available at \url{https://github.com/Ym-Shan/ORRC-SynA-natural-pruning}.

\end{abstract}


\section{Introduction}

\IEEEPARstart{A}{s} the inspiration for artificial neural networks (ANNs), the brain remains a focal point for both neuroscience and ANN research \cite{1}. Brain-inspired ANNs have achieved significant success in traditional and emerging multimodal research fields \cite{1,4}. Spiking Neural Networks (SNNs) have garnered attention due to their potential for energy efficiency and biological plausibility \cite{5}. Unlike ANNs, SNNs utilize event-based encoding, enabling asynchronous sparse input and approximating two-dimensional images through multiple time steps \cite{5,7}. This approach allows SNNs to perform comparably to ANNs, with the potential to outperform them due to the additional temporal correlation of event-encoded samples. SNNs have shown promise in various applications, including image classification \cite{8,9,10,11}, object detection \cite{12}, and speech recognition \cite{speech_recognition}. While there are multiple training methods for SNNs, including STDP unsupervised training and ANN-to-SNN conversion, direct training has emerged as an attractive option. However, this method faces challenges due to the presence of Dirac-Delta functions in the gradient of backpropagation. Neftci et al. proposed a surrogate gradient approach to address this issue \cite{18}, which has become the predominant method in current direct training approaches.
\IEEEpubidadjcol

In the pursuit of enhancing performance, researchers have explored distinct avenues. Some have focused on refining neuron efficiency through neurodynamic perspectives \cite{25}. Others have integrated attention mechanisms into SNNs, yielding notable improvements in model efficacy \cite{11,30}. Additionally, a cohort has pioneered the incorporation of residual learning principles into SNNs \cite{8,9,10}, enabling deeper training and directly amplifying network performance. Among these approaches, SEW-ResNet \cite{8} stands out as a prominent example of integrating residual structures into SNNs. Its significance extends beyond its own architecture, as it has been widely adopted in various advanced SNN models such as Spiking Transformers\cite{zhou2022spikformer} and SpikeGPT \cite{zhu2023spikegpt}, establishing itself as a fundamental work in the field of SNNs.
However, the conventional ResNet structure in SNNs, as exemplified by SEW-ResNet \cite{8}, faces a critical challenge. The addition operations in residual connections tend to reduce sparsity, leading to a deterioration in the overall model sparsity. This issue is particularly significant because the advantages of SNNs are primarily realized through hardware implementation, where sparsity plays a crucial role in energy efficiency.

With the advancement of SNN performance, power-efficient neuromorphic hardware has been developed \cite{davies2018loihi, akopyan2015truenorth}. These platforms excel at implementing spike-driven computations, allowing for the replacement of numerous Multiply-Accumulate (MAC) operations with a small number of Accumulate (AC) operations. This approach enables the system to bypass calculations for zero-valued positions, resulting in substantial energy savings. The efficiency stems from the fact that fixed-point adders are significantly superior to fixed-point multipliers in terms of both chip area and energy consumption.

To address the sparsity issue while mitigating performance degradation caused by high-level quantization with binary signals, we propose a novel approach that combines bitwise OR operations for residual connections with SynA attention. The bitwise OR operations retain more redundant information compared to traditional addition, while the SynA attention mechanism enhances the backbone's capability in extracting key features and reduces noise interference in feature extraction within the shortcuts. This integrated strategy not only improves model performance while maintaining a fully spike-driven architecture but also leads to an intriguing phenomenon: during our experiments, we observed that the firing rate of neurons in some or all shortcuts in the SynA-ResNet model occasionally drops to zero. This indicates that these shortcuts are not actively involved in the inference process. Remarkably, despite this apparent reduction in active connections, the model achieves higher accuracy. This characteristic offers dual benefits: it provides additional energy-saving advantages and enhances the potential for edge deployment of SynA-ResNet, making it particularly suitable for resource-constrained environments where both performance and efficiency are critical.

Our contributions can be summarized as follows:
\begin{enumerate}
\item{We propose a residual connection based on bitwise OR operation, which solves the problem of high energy consumption caused by SEW-ResNet's inability \cite{8} to achieve full AC operation and breaks the traditional concept that high quantization greatly affects model performance.}
\item{We design a set of attention module SynA for ORRC to propose a new training paradigm: low-level features from shortcuts complement high-level features rather than the same weight enhancement effect of traditional ADD residual connection. We also discover that the phenomenon of natural pruning for the first time.}
\item{Experiments validate the performance of our OR-Spiking ResNet and the efficacy of SynA, demonstrating the viability of bitwise residual connections. The proposed model achieves comparable or superior accuracy to other deep residual spiking neural networks while minimizing energy consumption across all datasets. This study pioneers the integration of attention mechanisms for significant energy reduction. We provide a comprehensive analysis of SynA's functionality, principles, synergistic effects, and the underlying causes of natural pruning, offering novel insights for edge device model deployment.}
\end{enumerate}

\section{Related Works}

\textbf{Residual connection in spiking neural networks.} To address the issues of gradient vanishing or degradation in deep-level SNNs, researchers have introduced residual learning into SNNs. Hu et al. \cite{31} first proposed a conversion method that scales continuous values in ANNs to correspond to the firing rate of SNNs, and introduced a compensation mechanism to offset conversion errors. Panda et al. \cite{32} proposed a backward residual connection with random softmax and validated it through transformation methods. Zheng et al. \cite{9} introduced tdBN, a normalization method enabling spiking residual networks to train up to 50 layers. Fang et al. \cite{8} proposed SEW-ResNet, which utilizes a unique network structure to train networks with over 100 layers. Hu et al. \cite{10} designed MS-ResNet, which considers the spiking communication specificity and neurodynamic characteristics, successfully training a network with 482 layers on the CIFAR-10 dataset.


\textbf{Attention in neural networks.} The advent of the CBAM module \cite{35} catalyzed widespread interest in attention mechanisms within the research community, owing to their exceptional performance and neuromorphic characteristics. In the realm of SNN research, Yao et al. \cite{36} pioneered the extension of attention to the temporal domain, considering inter-frame correlations in SNN event encoding. They subsequently broadened the application of channel and spatial attention in SNNs, culminating in the Multi-dimensional Attention (MA) module \cite{11}, which integrates temporal dimension attention. Concurrent research has yielded attention modules that address spatiotemporal correlations \cite{shan2024advancing, zhu2022tcja} and multiscale perspectives \cite{shan2024advancing}, achieving state-of-the-art accuracy through diverse approaches.


\textbf{Spike-driven in spiking neural networks.} Spike-driven computation in SNNs enables the substitution of numerous tensor multiplication operations with sparse addition operations, resulting in low-power characteristics. This capability necessitates the use of binary (0/1) tensors as operational units. While SEW-ResNet largely implements spike-driven operations \cite{8}, the first convolution after each shortcut involves non-binary input tensor elements, impeding full spike-driven implementation. Chen et al. \cite{38} addressed this issue by incorporating auxiliary accumulation paths during training, albeit at the cost of model performance and increased time steps. Conversely, Hu et al. \cite{10} designed MS-ResNet to ensure spike-driven behavior in all convolutional operations. The significance of spike-driven features in SNNs is gaining traction, with efforts to integrate them into spike transformers \cite{spikingformer_1}. It is anticipated that spike-driven computation will become a crucial link between SNNs and neuromorphic hardware.

\section{SynA-ResNet}
\label{sec:bit}
SynA-ResNet is an SNN model implemented using ORRC and equipped with the SynA attention module. In this section, we provide a comprehensive overview of the processing of static images and the network input format, which will be addressed in Section \ref{sec:bit/net}. Additionally, we delve into the LIF neuron model employed in our study, which will be discussed in Section \ref{sec:bit/lea}. Furthermore, we elaborate on the structure of the residual blocks in SynA-ResNet and provide a detailed explanation of the implementation of ORRC within these blocks, which will be covered in Section \ref{sec:bit/or}. The proposed SynA attention module will be thoroughly elucidated in Section \ref{sec:bit/spi}. Lastly, we introduce our novel approach, the bitwise operation residual spiking neural networks with SynA, in Section \ref{sec:bit/orrc}.

\subsection{Network input}
\label{sec:bit/net}

Previous studies \cite{kim2022rate} have shown that direct encoding \cite{wu2019direct} is more effective than rate encoding \cite{39} and time encoding \cite{40}. Therefore, we have chosen to utilize LIF neurons \cite{lif} as encoders in the initial module of ResNet (Conv-BN-LIF). Consequently, the convolutional layer preceding the block does not perform spike-driven computations. However, this has minimal impact since the first convolutional layer has the fewest parameters and is the least computationally complex. The additional complexity introduced by these non-spike-driven calculations is negligible. For static image datasets, we replicate the input images $T$ times and arrange them along the temporal dimension to capture event flow information. This results in a four-dimensional input array represented as $[T,C,H,W]$. The STBP algorithm is then employed to propagate the data forward and backward through the temporal and spatial dimensions.

\begin{figure*}[h]
  \centering 
  \includegraphics[width=\textwidth]{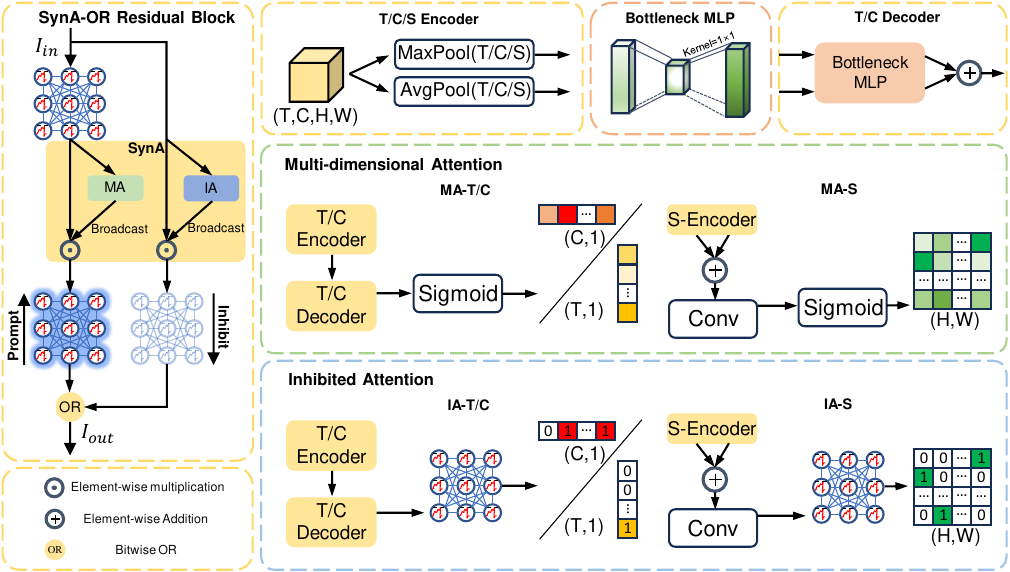}
  \caption{The structure and functional mechanism of the OR-Spiking ResNet residual block and SynA. The shading of the neurons represents the magnitude of their firing rate, with darker shades indicating higher firing rates. Similarly, the small squares representing attention weights use darker colors to denote greater attention weights.}
  \label{overview} 
\end{figure*}

\subsection{Leaky Integrate-and-Fire neuron model}
\label{sec:bit/lea}
The basic computing unit of neural networks is neuron, and in SNNs, and we generally use spiking neurons as the basic computing unit. Among these, the Leaky Integrate-and-Fire (LIF) neuron stands out as the optimal compromise between biological plausibility and computational efficiency. They are widely used in the fields of the brain like computing and computational neuroscience. The membrane behavior of postsynaptic neurons in the LIF model after receiving signals from presynaptic neurons before reaching the action potential release threshold can be described using this sub-threshold dynamics model:

\begin{equation}
\label{eq2}
\tau \frac{{d\boldsymbol{V}(t)}}{{dt}} =  - (\boldsymbol{V}(t) - {\boldsymbol{V}_{reset}}) + \boldsymbol{I}(t)
\end{equation}

where $\tau$ is a time constant, $\boldsymbol{V}(t)$ denotes the membrane potential of the postsynaptic neuron, $\boldsymbol{I}(t)$ is the input collected from presynaptic neurons, and $\boldsymbol{V}_{\text{reset}}$ is the reset potential which is set after activating the output spiking.

\begin{table}
  \centering
  \caption{hyper-parameter of LIF neuron}
  \label{table1}
  \begin{tabular}{cc}
    \toprule
    Hyper-parameter & Value \\
    \midrule
    $U_{\text{threshold}}  $ & 1.0  \\
    $U_{\text{reset}}      $ & 0  \\
    $\tau$                   & 2.0  \\
    surrogate function       & ATan()  \\
    membrane potential reset method                & Hard reset to 0  \\
    \bottomrule
  \end{tabular}

  \label{table1}
\end{table}

When a postsynaptic neuron receives a Spiking signal sent by a presynaptic neuron, the membrane potential of the postsynaptic neuron will increase. When it rises to $\boldsymbol{V}(t)>\boldsymbol{V}_{\text{threshold}}$, the postsynaptic neuron will emit a spike, and then $\boldsymbol{V}(t)$ will recover to the $ \boldsymbol{V}_{\text{reset}} $. For clearer reasoning and representation, a method similar to Fang et al. can be adopted \cite{8}, Decompose Equation \eqref{eq2} into an explicit iterative version as shown in Equation \eqref{eq3} to describe the integration, leakage, and firing of neurons, respectively.

\begin{equation}
\label{eq3}
\left\{ {\begin{array}{*{20}{l}}
{\boldsymbol{U}_t^n = \boldsymbol{H}_{t - 1}^n + \frac{1}{\tau }(\boldsymbol{I}_{t - 1}^n - (\boldsymbol{H}_{t - 1}^n - {\boldsymbol{U}_{reset}}))}\\
{\boldsymbol{S}_t^n = \Theta (\boldsymbol{U}_t^n - {\boldsymbol{U}_{threshold}})}\\
{\boldsymbol{H}_t^n = \boldsymbol{U}_t^n(1 - \boldsymbol{S}_t^n)}
\end{array}} \right.
\end{equation}

where $n$ and $t$ represent the layer index and time steps, respectively. $\boldsymbol{U}$ represents the neuron's membrane potential, $\boldsymbol{U}_{\text{t}}^{^\text{n}}$ represents the membrane potential of the neuron located in the $n$th layer at time step $t$. $\tau$ is still a time constant, $\boldsymbol{S}$ is a binary spiking tensor, $\boldsymbol{I}$ is the neuron input, $\Theta$ is the Heaviside step function, $\boldsymbol{H}$ represents the hidden state, $\boldsymbol{U}_{\text{reset}}$ is the reset potential of the neuron after spike, and $\boldsymbol{U}_{\text{threshold}}$ is the discharge threshold of the neuron.

In Equation \eqref{eq3}, The derivative of $\boldsymbol{S}_{\text{t}}^{^\text{n}}$ over $\boldsymbol{U}_{\text{t}}^{^\text{n}}$ is shown in Equation \eqref{eq4}:

\begin{equation}
\label{eq4}
\frac{{\partial \boldsymbol{S}_t^n}}{{\partial \boldsymbol{U}_t^n}} = \delta (\boldsymbol{U}_t^n - {\boldsymbol{U}_{threshold}})
\end{equation}

where $\delta$ denotes the Dirac-Delta function, its value tends to infinity at $\boldsymbol{U}_{\text{threshold}}$ and is always 0 at any other position, which leads to a dead neuron problem. In this work, we use the commonly used surrogate gradient method to solve this problem \cite{18}, which is to use the gradient of a certain function to replace the gradient of the Dirac-Delta function during backpropagation. The hyper-parameter settings for the LIF neuron model used are provided in Table \ref{table1}.

\subsection{OR-Spiking ResNet}
\label{sec:bit/or}
\begin{figure}[h]
	\begin{minipage}{0.32\linewidth}
		\vspace{3pt}
		\centerline{\includegraphics[width=\textwidth]{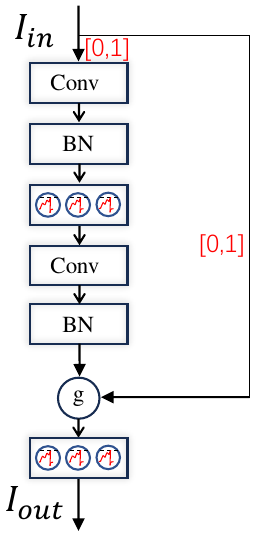}}
		\centerline{(a)}
	\end{minipage}
	\begin{minipage}{0.32\linewidth}
		\vspace{3pt}
		\centerline{\includegraphics[width=\textwidth]{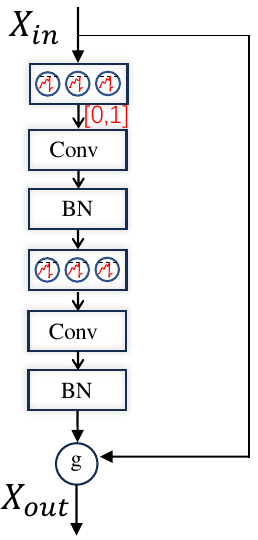}}
	 
		\centerline{(b)}
	\end{minipage}
	\begin{minipage}{0.32\linewidth}
		\vspace{3pt}
		\centerline{\includegraphics[width=\textwidth]{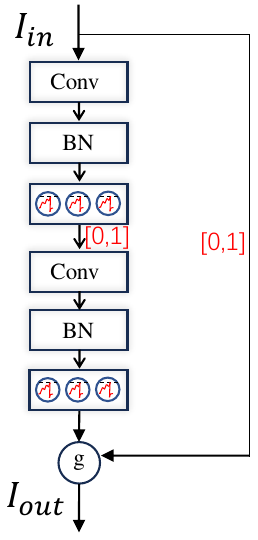}}
	 
		\centerline{(c)}
	\end{minipage}
 
	\caption{Mainstream Spiking ResNet network architecture. (a)Vanilla Spiking ResNet; (b)MS-ResNet; (c)SEW-ResNet, where Conv and BN represent convolution and Batch Normalization Layer, respectively. $g$ represents the element-wise functions of the aggregated backbone and shortcut.}
	\label{fig1}
\end{figure}

The residual block structure of our proposed OR-Spiking ResNet is illustrated in Figure \ref{overview}. Figure \ref{fig1} depicts the residual block structures of the Spiking ResNet variants proposed in previous works. The element-wise function computations for the three depicted sample block structures—Vanilla Spiking ResNet, MS-ResNet, and SEW-ResNet—are aligned with Equations \eqref{eq5}, \eqref{eq6}, and \eqref{eq7}, respectively. Notably, only the sample block structure of SEW-ResNet meets the criteria for bitwise element-wise operations, which require all operators to be binary. Consequently, following the precedent set by Fang et al., SEW-ResNet was selected as the standard for assessing the performance of Spiking ResNet \cite{8}. The execution of conventional addition operations alongside various bitwise element-wise functions within the SNN framework is summarized in Table \ref{table2}.

\begin{table}
\begin{center}
\caption{Bitwise element-wise functions}
\label{table2}
\begin{tabular}{ccc}
\toprule
Mode & Expression for g(x,y) & Code for g(x,y) \\
\midrule
ADD   & $x+y$                &  $x+y$              \\
AND   & $x \cap y$          &  $x \cdot y $    \\
IAND  & $(\neg x)\cap y $   &  $(1-x) \cdot y$  \\
OR    & $x \cup y$          &  $(x+y)-(x\cdot y)$\\
\bottomrule
\end{tabular}
\end{center}
\end{table}

\begin{equation}
\label{eq5}
\boldsymbol{I_{out}} = \mathcal{SN}(g(\boldsymbol{W}_{2} \cdot \mathcal{SN}(\boldsymbol{W}_{1}\cdot {\boldsymbol{I}_{in}}),  \boldsymbol{I}_{in})).
\end{equation}

\vspace{-10pt}
\begin{equation}
\label{eq6}
\boldsymbol{X_{out}} = g(\boldsymbol{W}_{2} \cdot \mathcal{SN}( \boldsymbol{W}_{1}\cdot \mathcal{SN}({\boldsymbol{I}_{in}})), \boldsymbol{I}_{in}).
\end{equation}

\vspace{-10pt}
\begin{equation}
\label{eq7}
\boldsymbol{I_{out}} = g(\mathcal{SN}(\boldsymbol{W}_{2} \cdot \mathcal{SN}(\boldsymbol{W}_{1}\cdot {\boldsymbol{I}_{in}})), \boldsymbol{I}_{in}).
\end{equation}

where $\mathcal{SN}(\cdot)$ is the spiking neuron. $\boldsymbol{W_{1}}$ and $\boldsymbol{W_{2}}$ denote the conv-BN weight of the first and second on the main road. And $g(\cdot)$ is the function between the main road and the branch chain, details of which can be seen in Table \ref{table2}.

The principal distinction between ORRC and the conventional ADD residual connection lies in the binarization of tensor elements within the outcomes of ORRC based on spike events. This implies that low-level feature maps from shortcuts merely supplement high-level ones without augmenting them. In contrast, the outcome tensor elements in ADD residual connections encompass three values (0, 1, and 2), indicating potential enhancement effects. While ORRC and ADD residual connections share a common response area, the results of ORRC lack the attribute of strength, introducing additional errors that lead to diminished network performance. Moreover, the binarization of tensor elements in the ORRC results addresses the issue of non-spike-driven operations in the first convolutional layer post-ORRC, thereby achieving fully spike-driven operations.

\IEEEpubidadjcol

\subsection{Synergistic Attention}
\label{sec:bit/spi}

The integration of attention serves as a potent mechanism augmenting a network's capacity to focus on specific elements within complex environments. This mechanism boasts biological plausibility, mirroring synaptic transmissions of electrical signals between neurons, thereby allowing precise regulation of neuronal communication by modulating synaptic weights \cite{28}.

As previously discussed, the distinctive attributes of ORRC and ADD residual connections give rise to distinct learning modes. In order to alleviate the additional errors caused by the lack of strength attributes in ORRC, we have designed the Synergistic Attention (SynA) module to assist ORRC in achieving high accuracy through an alternative learning mode. We have selected the proven-effective MA module \cite{11} as the facilitating component of our SynA module applied to the backbone. For the Inhibitory Attention (IA) component used in the shortcuts, we have leveraged the unique characteristics of spiking neurons (which do not fire when the membrane potential does not reach the threshold) to filter out noise in the shortcuts. This innovation builds upon the findings of Yao et al., who demonstrated the effectiveness of attention mechanisms in suppressing secondary features in the background \cite{11}. In contrast, our IA module takes noise suppression to the next level by leveraging the inherent limitations of neuron firing.

The structures of the MA and IA components are illustrated in Figure \ref{overview}, in the IA-T and IA-C modules, we first perform maximum pooling and average pooling operations on the temporal or spatial dimensions to obtain their average pooling features and maximum pooling features on the temporal and channel dimensions, respectively. Then, we extract the features through a bottleneck MLP and integrate them using LIF neurons to obtain attention weight tensors on the temporal or channel dimensions. This step can be represented as Equation \eqref{eq8} and Equation \eqref{eq9}.

\begin{equation}
\begin{aligned}
\label{eq8}
    \boldsymbol{W}_{T} = &\mathcal{SN}(\boldsymbol{W}^{m}_{t1}(\mathrm{ReLU}(\boldsymbol{W}^{n}_{t0}(\mathrm{AvgPool}(\boldsymbol{X_{in}}))))  \\&
    +\boldsymbol{W}^{m}_{t1}(\mathrm{ReLU}(\boldsymbol{W}^{n}_{t0}(\mathrm{MaxPool}(\boldsymbol{X_{in}}))))).
\end{aligned}
\end{equation}

where $\boldsymbol{W}_{T}$ is the temporal attention score. And $\boldsymbol{W}^{m}_{t1}\in\mathbb{R}^{T\times\frac{T}{r}}$, $\boldsymbol{W}^{n}_{t0} \in\mathbb{R}^{\frac{T}{r}\times T}$ are the weights of two shared dense layers. $\mathcal{SN}$ is the spiking neuron. $r$  is the temporal dimension reduction factor used to manage its computing overhead. 
\begin{equation}
\begin{aligned}
\label{eq9}
     \boldsymbol{W}_{C} = &\mathcal{SN}(\boldsymbol{W}^{m}_{c1}(\mathrm{ReLU}(\boldsymbol{W}^{n}_{c0}(\mathrm{AvgPool}(\boldsymbol{X_{in}}))))  \\&
    +\boldsymbol{W}^{m}_{c1}(\mathrm{ReLU}(\boldsymbol{W}^{n}_{c0}(\mathrm{MaxPool}(\boldsymbol{X_{in}}))))).
\end{aligned}
\end{equation}

where $\boldsymbol{W}_{C}$ is the channel attention score. And $\boldsymbol{W}^{m}_{c1}\in\mathbb{R}^{C\times\frac{C}{k}}$, $\boldsymbol{W}^{n}_{c0} \in\mathbb{R}^{\frac{C}{k}\times C}$ are the weights of two shared dense layers. $\mathcal{SN}$ is the spiking neuron. $k$  is the temporal dimension reduction factor used to manage its computing overhead.

Since IA-S, as shown in Figure \ref{overview}, is an attention module in the spatial dimension, it is not necessary or possible to use squeezes and extraction blocks. However, for better feature fusion, spatial addition has been modified to add on the channel dimension, and a convolutional layer with 2 input channels and 1 output channel has been added later. The attention weight calculation method in the spatial dimension is as follows:

\begin{equation}
\label{eq10}
\begin{split}
\boldsymbol{W_S} = \mathcal{SN}(f^{k \times k}[\mathrm{MaxPool}(\boldsymbol{X}_{in}); \mathrm{AvgPool}(\boldsymbol{X}_{in})])
\end{split}
\end{equation}

where $\boldsymbol{W}_{S}$ is the spatial attention score. And $f^{k \times k}$ is the 2D convolution operation.  $\mathrm{MaxPool}(\boldsymbol{X}_{in})$ and $\mathrm{AvgPool}(\boldsymbol{X}_{in})$ represent the results of global average-pooling and max-pooling layer respectively. $\mathcal{SN}$ is the spiking neuron. The attention mechanism of temporal, channel, and spatial dimensions and the combination of input are all simple tensor multiplication. Due to the different sizes of attention weights and inputs, attention weights need to be broadcasted (copied) according to order, so that they have the same size as the input. Then, the calculation shown in Equation \eqref{eq11} is applied to the input.

\begin{equation}
\label{eq11}
\begin{split}
\boldsymbol{X_{out}} = \boldsymbol{W_{k}} \cdot \boldsymbol{X_{in}}, k \in \left\{ {T, C, S} \right\}
\end{split}
\end{equation}

\subsection{OR-Spiking ResNet with Synergistic Attention}
\label{sec:bit/orrc}

As previously mentioned, we selected the bitwise operation OR, which exhibits the highest redundancy, as the method for residual connection. Subsequently, we enhanced its primary features through ORRC while suppressing background noise from being extracted as features. This approach enables the training of spike-driven models with high performance using a distinctly different training paradigm.

The mechanism and desired effects of SynA in OR-Spiking ResNet are illustrated in Figure \ref{overview} (please note the glow and blurring effects). To evaluate its effectiveness in OR-Spiking ResNet, we conducted comprehensive ablation experiments in Section \ref{sec:exp/abl/res} and provided a detailed analysis of the energy efficiency source of SynA in Section \ref{sec:exp/abl/ana}. During the experimental validation of SynA's effectiveness, we observed that as the residual model with SynA continued training, some or all of the shortcuts naturally dropped out without any performance loss. We refer to this phenomenon as "natural pruning." A thorough analysis of this intriguing feature, as well as the impact and underlying principle of attention, will be conducted in the ablation research section of Section \ref{sec:exp/abl/ana}.

\section{Analysis of energy consumption}
\label{sec:ana}
The energy efficiency of SNNs stems from the ability to substitute fewer AC operations for MAC operations in spike-driven computations, with each AC operation consuming less energy than a MAC operation. Analyzing energy consumption in SNNs that don't employ rate coding or time coding is more intricate. This is due to the initial module's convolutional operation, acting as an encoder, which performs MAC operations that don't align with spike-driven calculations. In contrast, other convolutional operations execute AC and MAC operations separately, contingent on whether the input is even encoded. Additionally, we incorporate spike count to compute the total energy consumption ($\boldsymbol{E}$), calculated using MAC ($\boldsymbol{N}_{\text{MAC}}$) and AC ($\boldsymbol{N}_{\text{AC}}$) operations, as a metric for assessing model performance.

The quantification of MAC and AC operations hinges on factors such as firing rate and the nature of operations in each module. This encompasses convolution operations, batch normalization, linear transformations, and operations within various attention modules (encompassing activation functions in the attention model, among others). Aligning with established practice in prior SNNs studies \cite{10,11,14}, we assume that all models operate on hardware featuring a 45nm process, where $\boldsymbol{N}_{\text{MAC}}=4.6pJ$ and $\boldsymbol{N}_{\text{AC}}=0.9pJ$. The calculation formula for $\boldsymbol{E}$ is as follows:
\begin{equation}
\label{eq12}
    \begin{aligned}
\boldsymbol{E}_{total}&=\boldsymbol{E}_{MAC}\cdot \boldsymbol{FL}_{conv}^1+\\&\boldsymbol{E}_{AC}\cdot (\sum_{n=2}^N \boldsymbol{FL}_{conv}^n \cdot fr^{n} +\sum_{m=1}^M \boldsymbol{FL}_{fc}^m \cdot fr^{m}).
\end{aligned}
\end{equation}

where $\boldsymbol{N}$ and $\boldsymbol{M}$ are the total number of Conv and FC layers, $\boldsymbol{E}_{MAC}$ and $\boldsymbol{E}_{AC}$ are the energy costs of MAC and AC operations, and $fr^{m}$, $fr^{n}$, $\boldsymbol{FL}_{conv}^n$ and $\boldsymbol{FL}_{fc}^m$ are the firing rate and FLOPs of the $n$-th Conv and $m$-th FC layer.

\section{Experiments}
\label{sec:exp}

\begin{table*}[]
\caption{Comparison with previous works on neuromorphic dataset}
\label{table4}
\centering
\begin{tabular}{@{}cccccccc@{}}
\toprule
\multirow{2}{*}{Datasets}     & \multirow{2}{*}{\begin{tabular}[c]{@{}c@{}}Method\\ $\checkmark$ Rsepresents Spike-driven SNN\end{tabular}}      & \multirow{2}{*}{\begin{tabular}[c]{@{}c@{}}Spike\\ Number(K)\end{tabular}} & \multicolumn{2}{c}{Calculation(M)} & \multirow{2}{*}{\begin{tabular}[c]{@{}c@{}}Energy\\ Consumption(mJ)\end{tabular}} & \multirow{2}{*}{Top-1 Acc.(\%)} & \multirow{2}{*}{Pruning} \\ \cmidrule(lr){4-5}
                              &                              &                                                                            & MAC          & AC                  &                                                                                   &                                 &                          \\ \midrule
\multirow{15}{*}{DVS Gesture} & SLAYER\cite{53}              & -                                                                          & -            & -                   & -                                                                                 & 93.64                           &                          \\
                              & Rollout\cite{54}             & -                                                                          & -            & -                   & -                                                                                 & 97.20                            &                          \\
                              & LIAF-NET\cite{55}            & -                                                                          & -            & -                   & -                                                                                 & 97.60                            &                          \\
                              & tdBN\cite{9}                 & -                                                                          & -            & -                   & -                                                                                 & 96.90                            &                          \\
                              & PLIF\cite{56}                & -                                                                          & -            & -                   & -                                                                                 & 96.90                            &                          \\
                              & STS-ResNet\cite{57}          & -                                                                          & -            & -                   & -                                                                                 & 96.70                            &                          \\
                              & Vanilla Spiking ResNet\cite{9}        & 703.04                                                                     & 4055.89      & 808.62             & 19.393                                                                             & 96.18                           &                          \\
                              & MS-ResNet\cite{10}\checkmark  & 1712.88                                                                      & 840.43       & 1104.73              & 4.868                                                                               & 95.49                            &                          \\
                              & SEW-ResNet\cite{8}            & 1038.72                                                                    & 4462.48      & 844.50              & 21.296                                                                             & 96.18                           &                          \\
                              & AND-SEW-ResNet\cite{8}        & 947.39                                                                     & 840.43       & 986.95              & 4.762                                                                              & 97.22                           &                          \\
                              & IAND-SEW-ResNet\cite{8}       & 1049.13                                                                   & 840.43       & 1163.89              & 4.922                                                                              & 96.88                           &                          \\ \cmidrule(l){2-8} 
                              & OR-Spiking ResNet\checkmark   & 928.62                                                                     & 840.43       & 1000.19              & 4.774                                                                              & 95.83                            &                          \\
                              & SynA-T-ResNet\checkmark & \textbf{646.09}                                                            & \textbf{839.91}       & \textbf{745.68}     & \textbf{4.543}                                                                     & 97.57                  & \textbf{shortcut 2}               \\
                              & SynA-C-ResNet\checkmark & \textbf{608.15}                                                            & \textbf{838.77}       & \textbf{777.22}      & \textbf{4.567}                                                                     & \textbf{97.92}                  & \textbf{all shortcuts}   \\
                              & SynA-S-ResNet\checkmark & 766.95                                                           & \textbf{840.38}       & 837.03      & \textbf{4.626}                                                                     & 96.18                  & \textbf{shortcut 3}   \\ \midrule
\multirow{16}{*}{CIFAR10-DVS} & DART\cite{58}                & -                                                                          & -            & -                   & -                                                                                 & 65.78                           &                          \\
                              & Rollout-ANN\cite{59}         & -                                                                          & -            & -                   & -                                                                                 & 66.75                           &                          \\
                              & Rollout\cite{54}             & -                                                                          & -            & -                   & -                                                                                 & 66.80                            &                          \\
                              & LIAF-NET\cite{55}            & -                                                                          & -            & -                   & -                                                                                 & 70.40                            &                          \\
                              & tdBN\cite{9}                 & -                                                                          & -            & -                   & -                                                                                 & 67.80                            &                          \\
                              & PLIF\cite{56}                & -                                                                          & -            & -                   & -                                                                                 & 67.80                            &                          \\
                              & STS-ResNet\cite{57}          & -                                                                          & -            & -                   & -                                                                                 & 69.20                            &                          \\
                              & Vanilla Spiking ResNet\cite{9}        & \textbf{76.24}                                                                     & 274.68      & \textbf{87.54}             & 1.346                                                                             & 75.78                           &                          \\
                              & MS-ResNet\cite{10}\checkmark  & 131.05                                                                     & 11.92       & 125.43              & 0.172                                                                              & 76.56                              &                          \\
                              & SEW-ResNet\cite{8}            & 117.31                                                                     & 332.64      & 87.89              & 1.613                                                                              & 72.85                           &                          \\
                              & AND-SEW-ResNet\cite{8}        & 144.99                                                                     & 11.92       & 130.04              & 0.176                                                                              & 78.12                          &                          \\
                              & IAND-SEW-ResNet\cite{8}       & 139.58                                                                     & 11.92       & 115.66              & 0.163                                                                              & 77.34                          &                          \\ \cmidrule(l){2-8} 
                              & OR-Spiking ResNet\checkmark   & 123.56                                                            & 11.92       & 124.66      & 0.171                                                                     & 69.92                  &                          \\ 
                              & SynA-T-ResNet\checkmark & 116.06                                                            & \textbf{11.78}       & 122.45     & 0.168                                                                     & \textbf{78.51}                  & \textbf{all shortcuts}               \\
                              & SynA-C-ResNet\checkmark & 102.44                                                            & 12.04       & 105.06      & \textbf{0.153}                                                                     & 74.41                  & \textbf{shortcut 2}   \\
                              & SynA-S-ResNet\checkmark & 109.77                                                            & \textbf{11.98}       & 113.92      & \textbf{0.161}                                                                     & 75.00                  & \textbf{all shortcuts}   \\ \midrule
\end{tabular}
\end{table*}

In this section, we commence by conducting a comprehensive investigation and comparative analysis of the performance of SynA-ResNet, incorporating SynA modules of varying dimensions, across a diverse range of widely-utilized datasets. Subsequently, we provide a detailed elucidation of the methodology employed in determining the optimal placement of the attention mechanism within the network architecture. Furthermore, we undertake a series of experiments involving the incremental addition or removal of modules to rigorously validate the efficacy and robustness of the SynA approach. Finally, we present an in-depth discussion, leveraging both visualization techniques and quantitative analysis, to shed light on the underlying factors contributing to the low power consumption of the proposed model and the intriguing phenomenon of natural pruning that emerges during the training process.

To better illustrate the feasibility of ORRC and SynA while collecting energy consumption data, we implemented the network structure in \cite{8,9,10} using the PyTorch and SpikingJelly \cite{45} frameworks with standard hyper-parameters on each dataset, and compared performance at a depth of 21 layers.

\subsection{Performance evaluation in classification tasks}
\label{sec:exp/per}
In this section, we will evaluate the effectiveness and performance of our proposed ORRC and SynA-ResNet on two event-encoded neuromorphic datasets, the DVS Gesture dataset, CIFAR10-DVS dataset, and two static image datasets, MNIST and Fashion-MNIST. For specific experimental parameter settings, please refer to Section \ref{sec:experimental}.

\subsubsection{Experiment on neuromorphic datasets}

The DVS Gesture gesture recognition dataset is an event-based dataset that contains a sequence of 11 gestures \cite{46}. It has 1176 training set samples and 288 test set samples, and the training set and training set are shot by 23 and 6 subjects under three lighting conditions, respectively. We encode the event sequence of DVS Gesture at T=32 into the format of [Time Step, Channel, Height, Width] for our training and testing sets, with each frame of the image having a size of $128\times 128$ pixels.

The CIFAR10-DVS dataset \cite{47} is a conversion from the CIFAR10 dataset, consisting of ten categories totaling 10000 images. The size of each frame has also been expanded to $128\times 128$ pixels. Due to the additional undesired artifacts included in the neuromorphic datasets obtained based on the conversion method, therefore, how to distinguish the categories of targets in complex backgrounds with additional errors is a challenging recognition task.

As illustrated in Table \ref{table4}, the OR-Spiking ResNet architecture demonstrated remarkable energy efficiency, consuming a mere 4.774mJ (T=32) to perform image classification on the DVS Gesture dataset. When compared to the representative Vanilla Spiking ResNet and SEW-ResNet models, which employ ADD residual connections, our approach achieved a significant reduction in computational complexity. Specifically, we replaced the 3215.46M MAC operations in Vanilla Spiking ResNet with 191.57M MAC operations and the 3622.05M MAC operations in SEW-ResNet with 155.69M AC operations. This optimization resulted in a 3.06-fold and 3.46-fold decrease in the model's energy consumption, respectively. Furthermore, the incorporation of SynA-T, SynA-C, or SynA-S modules led to a notable improvement in classification accuracy while simultaneously lowering energy consumption. On the DVS Gesture dataset, the OR-Spiking ResNet model augmented with SynA-C achieved a 1.74\% accuracy improvement compared to SEW-ResNet using ADD residual connections. Remarkably, this enhancement was attained using only 608.15K spikes (0.8 spikes per neuron), a reduction of 430.57K spikes compared to SEW-ResNet, resulting in further diminished power consumption. Moreover, despite the OR-Spiking ResNet architecture having an additional 5 neuron layers compared to Vanilla Spiking ResNet and an extra neuron layer compared to MS-ResNet, our model consistently required the fewest spikes to accomplish gesture classification tasks, underscoring its exceptional efficiency.

\begin{table*}[]
\caption{Comparison with previous works on MNIST and Fashion-MNIST}
\label{table5}
\centering
\begin{tabular}{@{}cccccccc@{}}
\toprule
\multirow{2}{*}{Datasets}      & \multirow{2}{*}{\begin{tabular}[c]{@{}c@{}}Method\\ $\checkmark$ Rsepresents Spike-driven SNN\end{tabular}}      & \multirow{2}{*}{\begin{tabular}[c]{@{}c@{}}Spike\\ Number(K)\end{tabular}} & \multicolumn{2}{c}{Calculation(M)} & \multirow{2}{*}{\begin{tabular}[c]{@{}c@{}}Energy\\ Consumption(mJ)\end{tabular}} & \multirow{2}{*}{Top-1 Acc.(\%)} & \multirow{2}{*}{Pruning} \\ \cmidrule(lr){4-5}
                               &                              &                                                                            & MAC          & AC                  &                                                                                   &                                 &                          \\ \midrule
\multirow{8}{*}{MNIST}         & Vanilla Spiking ResNet\cite{9}                & 372.07                                                                      & 1323.6       & 346.69               & 6.405                                                                              & 99.62                           &                          \\
                               & MS-ResNet\cite{10}\checkmark          & 564.54                                                                      & 14.10         & 579.45               & 0.590                                                                              & 99.54                           &                          \\
                               & SEW-ResNet\cite{8}                    & 420.62                                                                      & 1542.19       & 289.26               & 7.359                                                                              & 99.56                           &                          \\
                               & AND-SEW-ResNet\cite{8}                & 439.62                                                                      & 14.10         & 392.94               & 0.423                                                                              & 99.55                            &                          \\
                               & IAND-SEW-ResNet\cite{8}               & 539.75                                                                      & 14.10         & 410.68                & 0.439                                                                              & 99.59                           &                          \\ \cmidrule(l){2-8} 
                               & OR-Spiking ResNet\checkmark           & 494.12                                                                      & 14.10         & 444.11               & 0.469                                                                              & 99.51                           & \textbf{}                \\
                               & SynA-T-ResNet\checkmark & \textbf{239.98}                                                             & \textbf{13.37}         & \textbf{221.54}       & \textbf{0.265}                                                                      & \textbf{99.63}                  & \textbf{all shortcuts}               \\
                               & SynA-C-ResNet\checkmark & \textbf{247.29}                                                              & \textbf{13.95}         & \textbf{250.65}       & \textbf{0.293}                                                                      & 99.60                  & \textbf{shortcut 1}             \\ \midrule
\multirow{9}{*}{Fashion-MNIST} & Vanilla Spiking ResNet\cite{9}                & 566.81                                                                       & 1323.6       & 463.12                & 6.509                                                                              & 93.94                           &                          \\
                               & MS-ResNet\cite{10}\checkmark          & 760.04                                                                      & 14.10         & 610.19                & 0.618                                                                              & 93.66                           &                          \\
                               & SEW-ResNet\cite{8}                    & 627.18                                                                      & 1542.19       & 393.78                & 7.453                                                                              & 94.15                           &                          \\
                               & AND-SEW-ResNet\cite{8}                & 728,34                                                                       & 14.10         & 578.24               & 0.589                                                                              & 94.18                           &                          \\
                               & IAND-SEW-ResNet\cite{8}               & 857.13                                                                         & 14.10         & 590.05               & 0.600                                                                              & 94.27                           &                          \\ \cmidrule(l){2-8} 
                               & OR-Spiking ResNet\checkmark           & 691.90                                                                      & 14.10         & 603.31                & 0.612                                                                              & 94.21                           &                          \\
                               & SynA-T-ResNet\checkmark & \textbf{531.50}                                                             & \textbf{13.50}         & \textbf{385.84}                & \textbf{0.413}                                                                     & \textbf{94.43}                  & \textbf{shortcut 1,2}   \\
                               & SynA-C-ResNet\checkmark & \textbf{504.52}                                                             & \textbf{13.95}         & 441.87                & \textbf{0.465}                                                                     & 93.98                  & \textbf{shortcut 1}   \\
                               & SynA-S-ResNet\checkmark & \textbf{368.24}                                                             & \textbf{13.53}         & \textbf{220.50}                & \textbf{0.264}                                                                     & 93.48                  & \textbf{all shortcuts}   \\ \bottomrule
\end{tabular}
\end{table*}

Similarly, on the CIFAR10-DVS dataset, our OR-Spiking ResNet model demonstrated a significant reduction in computational complexity by replacing the 262.76M MAC operations of the Vanilla Spiking ResNet and the 320.72M MAC operations of the SEW-ResNet with a mere 37.12M and 36.77M AC operations, respectively. This optimization resulted in a substantial decrease in inference power consumption, yielding a 7.87-fold reduction compared to the Vanilla Spiking ResNet and a 9.43-fold reduction compared to the SEW-ResNet. Moreover, the incorporation of SynA modules, regardless of their dimensionality, led to a notable enhancement in the model's classification performance, with improvements ranging from 4.49\% to 8.59\%. These results provide compelling evidence that our SynA mechanism is indeed capable of extracting salient information from the extensive redundant data preserved by the ORRC encoding scheme, thereby significantly boosting the model's discriminative power while maintaining a highly efficient computational profile. In terms of the number of spikes, due to the presence of substantial noise in the CIFAR10-DVS dataset and the redundant information introduced by ORRC, SynA-ResNet requires more spikes during the inference process compared to Vanilla Spiking ResNet, which has fewer neurons, even after SynA refines the redundant information. However, when compared to residual SNN models with the same number of neurons, SynA-ResNet necessitates fewer spikes for inference.

\subsubsection{MNIST and Fashion-MNIST classification}

The MNIST dataset \cite{50} was handwritten by National Institute of Standards and Technology (NIST) employees and high school students with the ten digits 0-9, which were then normalized and anti-aliased. It is a model of deep learning datasets, consisting of 60000 training images and 10000 test images with a size of $28\times 28$ pixels.

The Fashion-MNIST dataset \cite{51} is a clothing image dataset produced by the German company Zalando based on products. Its sample format and quantity are identical to MNIST, and its emergence solves the problem of mainstream work achieving almost perfect results on the MNIST dataset, resulting in a lack of evaluation indicators for the performance of deep learning models.

As shown in Table \ref{table5}, on the MNIST dataset with a long history and relatively simple tasks, our OR-Spiking ResNet achieved similar results to previous SOTA models. After being equipped with SynA-T, the classification accuracy of the network increased by 0.12\%, while the number of spikes required to classify an image decreased by approximately 51.43\%, and the energy consumption required to classify an image decreased by approximately 43.5\%; If equipped with SynA-C, the classification accuracy of the model will be improved by 0.09\%, while the number of spikes required to classify an image will be reduced by twice, and the energy consumption will be reduced by 0.5 times. This indicates that OR-Spiking ResNet has achieved almost perfect classification accuracy on the MNIST dataset, while our SynA can only perform minor optimizations in terms of energy consumption. When the model accuracy approaches saturation, the effect of attention will shift from optimizing the model classification accuracy to significantly reducing model energy consumption. Our model can also achieve similar or even higher accuracy to other mainstream models on the Fashion-MNIST dataset with low energy consumption. The above experiment proves that the performance of ORRC is not significantly lower than that of ADD residual connection in simple tasks. On the MNIST and FashionMNIST datasets, we also observed the phenomenon of 'natural pruning' in OR Spiking ResNet with SynA added, indicating that the 'natural pruning' phenomenon is not only applicable to neuromorphic datasets.

\subsection{Ablation study}
\label{sec:exp/abl}
In Section \ref{sec:exp/per}, we demonstrated and analyzed the competitive performance of our OR-Spiking ResNet, which is equipped with SynA and uses ORRC, on multiple mainstream datasets. Below, we conducted a detailed ablation study on the selection of SynA's action mode in the network, the source of model performance, and other aspects. We demonstrated the process of selecting the location of SynA in the network in Section \ref{sec:exp/abl/syn}; In Section \ref{sec:exp/abl/res}, a reasonable analysis was conducted on the reasons for the effectiveness of SynA through the spiking response of OR-spiking ResNet and the ablation experiment of SynA; In section \ref{sec:exp/abl/ana}, the reason for natural pruning is explained by visualizing the trend of firing rate of each layer of neurons in the model over time and epoch, and the attention distribution in the network can be reasonably inferred from this.

\subsubsection{SynA location in OR-Spiking ResNet}
\label{sec:exp/abl/syn}

\begin{table*}[]
\caption{The first OR-SEW module structure of OR-Spiking ResNet}
\label{B-1-table-1}
\centering
\begin{tabular}{@{}ccccccccccccc@{}}
\toprule
SynA loaction & Backbone                  & Residual Block     & Backbone after Residual Block                   \\ \midrule
a             & c128k3s2p1-BN-MA-LIF-c128k3s1p1-BN-LIF & c128k1s2-BN-IA-LIF & c128k3s1p1-BN-MA-LIF-c128k3s1p1-BN-LIF \\
b             & c128k3s2p1-BN-LIF-c128k3s1p1-BN-MA-LIF & c128k1s2-BN-IA-LIF & c128k3s1p1-BN-LIF-c128k3s1p1-BN-MA-LIF \\
c             & c128k3s2p1-BN-MA-LIF-c128k3s1p1-BN-LIF & c128k1s2-BN-IA-LIF & c128k3s1p1-BN-LIF-c128k3s1p1-BN-LIF    \\
d             & c128k3s2p1-BN-LIF-c128k3s1p1-BN-MA-LIF & c128k1s2-BN-IA-LIF & c128k3s1p1-BN-LIF-c128k3s1p1-BN-LIF    \\ \bottomrule
\end{tabular}
\end{table*}

\begin{table*}[]
\caption{Ablation study on the Dimension and Position of SynA Action}
\label{B-1-table-2}
\centering
\begin{tabular}{@{}ccccccccccccc@{}}
\toprule
              & a-T   & b-T   & c-T   & d-T   & a-C   & b-C   & c-C   & d-C   & a-S    & b-S   & c-S    & d-S   \\ \midrule
DVS Gesture   & -0.35 & +1.74 & +1.74 & \textbf{+1.74} & +0.69 & \textbf{+2.09} & 0.00     & +0.35 & -4.20   & -0.35 & \textbf{+0.35}   & -0.70 \\
CIFAR10 DVS   & \textbf{+8.59} & +8.4  & +6.83 & +6.25 & +4.29 & \textbf{+4.49} & +1.75 & +3.32 & -56.84 & +0.78 & -53.52 & \textbf{+5.08} \\
MNIST         & \textbf{+0.12} & +0.04 & 0.00  & 0.00 & +0.04 & \textbf{+0.09} & -0.03 & +0.01 & -87.99  & -88.03 & -84.41  & -88.04      \\
Fashion-MNIST & -0.78 & -0.96 & \textbf{+0.22} & -0.38 & \textbf{-0.23} & -0.25 & -0.01 & -0.15 & \textbf{-0.73} & -0.82 & -84.11 & -66.5 \\ \bottomrule
\end{tabular}
\end{table*}

\IEEEpubidadjcol

Due to the existence of maximum pooling operation in the calculation of attention weights in SynA, considering that maximum and minimum values can cause feature loss, high sensitivity resulting in overfitting, and insensitivity to slight shifts in maximum or minimum values, resulting in reduced robustness, we only consider the way the MA module in SynA operates after the BN layer.

In Table \ref{B-1-table-1}, we present all potential operating modes of SynA within the OR-SEW module, and the impacts of each mode on time, channel, and spatial dimensions are detailed in Table \ref{B-1-table-2}. For each dataset we considered, we selected the mode of operation that yielded the highest classification accuracy in each dimension and provided the corresponding results in Tables \ref{table4} and \ref{table5}, respectively (in the table, "a-T" signifies T attention with "a" as the mode of operation). 

Preceding the first OR-SEW module, there is also a feature extraction module, encompassing the second through fifth convolutional layers of OR-spiking ResNet21. For this module, the insertion position is determined by the specific location of SynA's action, adopting the same insertion position as the MA module in the backbone of the corresponding Residual Block.

\subsubsection{Research on synergistic effects of SynA}
\label{sec:exp/abl/res}

\begin{figure}[h]
  \centering 
  \includegraphics[width=0.45\textwidth]{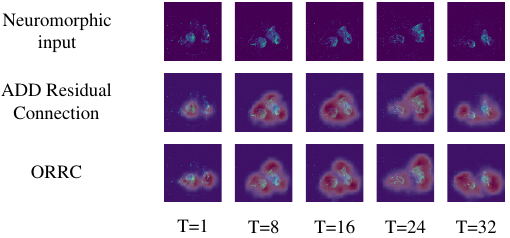}
  \caption{Characteristic strength heatmap for two types of residual connections. We take samples with T=1, 8, 16, 24, and 32 and visualize them when the DVS Gesture encoding is T=32, The purple in the figure is the background, the blue part is the target in the input sample, and the color depth in the red part indicates the Spiking intensity emitted by the LIF neurons at different positions output by the first shortcut in the network.}
  \label{fig5} 
\end{figure}

\begin{figure}[h]
  \centering 
  \includegraphics[width=0.45\textwidth]{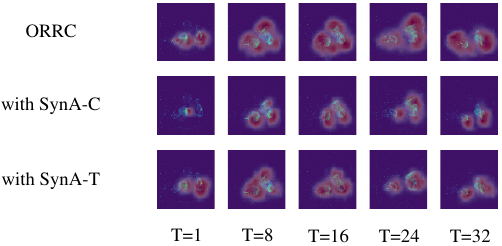}
  \caption{Characteristic strength heatmap of the SynA effect in both channel and temporal dimensions. }
  \label{fig6} 
\end{figure}

\begin{table}
\begin{center}
\caption{attention ablation study on OR-Spiking ResNet}
\label{table3}
\begin{tabular}{cc}
\toprule
Method & Top-1 Acc.(\%)  \\
\midrule
OR-Spiking ResNet                             & 95.83    \\
OR-Spiking ResNet with MA-T                   & 94.44             \\
OR-Spiking ResNet with IA-T                   & 96.18     \\
OR-Spiking ResNet with SynA-T                 & 97.57          \\
OR-Spiking ResNet with MA-C                   & 97.92    \\
OR-Spiking ResNet with IA-C                   & 95.49    \\
OR-Spiking ResNet with SynA-C                 & 97.92    \\
\bottomrule
\end{tabular}
\end{center}
\end{table}

\begin{table}
\begin{center}
\caption{Performance comparison between natural pruning and pruning before training}
\label{B-3-table-1}
\begin{tabular}{cc}
\toprule
Method                                          & Top-1 Acc.(\%) \\ \midrule
OR-Spiking ResNet with SynA-T                 & 97.57                 \\ 
OR-Spiking ResNet with SynA-C                 & 97.92                 \\ 
OR-Spiking ResNet with SynA-T no shortcut 2 & 95.83                    \\ 
OR-Spiking ResNet with SynA-C no shortcut 1,2,3 & 96.53                 \\ \bottomrule
\end{tabular}
\end{center}
\end{table}

\begin{figure*}[h]
  \centering 
  \includegraphics[width=1\textwidth]{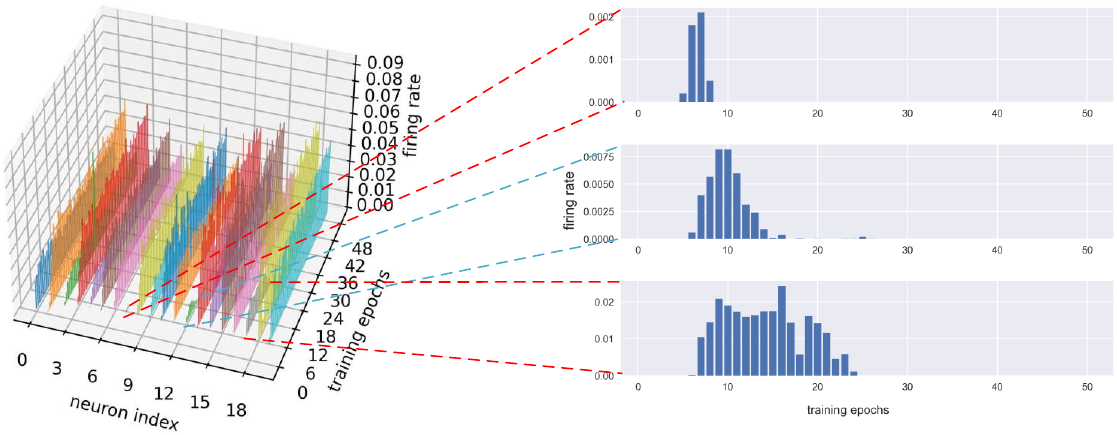}
  \caption{Thermogram of the changes in firing rate of all neurons in OR-Spiking ResNet with SynA-C added as training rounds increase. For a clearer observation, We created three bar charts to illustrate the changes in the firing rate of LIF neurons in shortcuts as the number of training rounds increased.}
  \label{fig7} 
\end{figure*}

\IEEEpubidadjcol

As demonstrated in Table \ref{table4}, our meticulously designed OR-Spiking ResNet architecture achieved a remarkable action classification accuracy of 95.83\% on the Dvs Gesture dataset. Building upon this foundation, we will conduct a comprehensive series of ablation experiments to thoroughly investigate and analyze the synergistic effects of the MA and IA modules within the SynA framework on the performance of OR-Spiking ResNet. Through these rigorous experiments, we aim to gain deeper insights into the intricate interplay between these critical components and their collective impact on the model's classification capabilities. By carefully examining the results of these ablation studies, we seek to unravel the underlying mechanisms that contribute to the exceptional performance of our proposed architecture and provide valuable guidance for future research endeavors in this domain.

As shown in Table \ref{table3}, in accordance with the optimal approach determined in Section \ref{sec:exp/abl/syn}, we incorporated the MA-T module into OR-Spiking ResNet. We observed that the MA-T module reduced the classification accuracy of the model by 1.39\%, possibly because it led the model to focus too much on less relevant temporal inputs. Next, we attempted to add only the IA module proposed in Section \ref{sec:bit/spi} to OR-Spiking ResNet to verify whether the performance improvement brought by SynA was solely attributed to the IA module. Through experiments, we find that the IA module applied in the temporal dimension did indeed enhance the classification accuracy of OR-Spiking ResNet, but it only increased by 0.35\%. This slight accuracy gain was a result of the IA-T module directing the model's attention toward less irrelevant information. However, when we applied MA-T and IA-T to OR-Spiking ResNet simultaneously, the model's classification performance on the Dvs Gesture dataset improved to 97.57\%. This marked an increase of 1.39\% compared to adding only IA-T. This demonstrates the synergy of SynA in the temporal dimension, indicating that the MA and IA modules complement each other. By pruning unwanted features in the shortcut while enhancing the backbone features, higher accuracy can be achieved, even though one of them may have a negative impact when used in isolation. Similarly, we improved the model's classification accuracy by 2.09\% by adding MA-C to OR-Spiking ResNet. When only IA-C was added, the model's classification accuracy decreased by 0.34\%. While OR-Spiking ResNet equipped with SynA-C did not exhibit a significant increase in classification accuracy compared to adding only MA-C, it introduced additional natural pruning capabilities. Moreover, it enabled pruning all three shortcuts, effectively transforming OR-Spiking ResNet into Spiking VGG, which aids in the deployment of the model on edge devices.

The following provides a visual analysis of the operational characteristics of ORRC and elucidates how SynA effectively enhances model classification accuracy. Figure \ref{fig5} presents a heatmap generated by visualizing the output tensor after the model executed the first ORRC operation. The observed phenomenon arises from the complex and confounding feature intensities introduced by the bitwise OR operation. This confusion is exponentially amplified when multiple ORRC operations are stacked, causing the network to focus on features with medium strengths. Consequently, the performance of ORRC is slightly lower than that of the ADD residual connection, as demonstrated by the detailed comparative experiments conducted in Section \ref{sec:exp/per}.

\begin{table*}[]
\caption{The network structure used by our method on various datasets}
\label{C-table-1}
\centering
\begin{tabular}{@{}cccccccc@{}}
\toprule
Dataset             & Network                                                                                                              \\ \midrule
\multirow{2}{*}{DVS Gesture}         & c64k7s2p3-BN-LIF-MPk3s2p1-\{c64k3s1p1-BN-LIF\}*4-(OR-SEW Block(c128))-(OR-SEW Block(c256))-\\
                    &(OR-SEW Block(c512))-AP-FC11                \\ \midrule
\multirow{2}{*}{CIFAR10-DVS}         & AdaptiveAP(48)-c64k3s2p1-BN-LIF-MPk3s2p1-\{c64k3s1p1-BN-LIF\}*4-(OR-SEW Block(c128))-(OR-SEW Block(c256))- \\ 
                    &      (OR-SEW Block(c512))-AP-FC10 \\ \midrule
\multirow{2}{*}{MNIST/Fashion-MNIST} & c64k3s1p1-BN-LIF-\{c64k3s1p1-BN-LIF\}*4-(OR-SEW Block(c128))-(OR-SEW Block(c256))- \\
                    & (OR-SEW Block(c512))-AP-FC10                         \\ \bottomrule
\end{tabular}
\end{table*}

The heatmap in Figure \ref{fig6}, visualized in the same manner and at the same position as in Figure \ref{fig5}, reveals a significant reduction in the size of the red areas. In Figure \ref{fig6}, the distribution of the red areas has evolved from surrounding the entire fist or joint to being concentrated on a portion of the fist or joint. This indicates that with the addition of SynA attention, the network has learned the critical factors for determining action categories – the motion trajectories of key parts such as joints. This strongly demonstrates that our SynA attention not only helps reduce the enlarged attention areas caused by ORRC but also focuses the network's attention on the most critical positions to learn essential temporal information. This explains why SynA attention does not perform well on static datasets like MNIST, as static images lack relevant temporal information.

\subsubsection{Analysis of natural pruning}
\label{sec:exp/abl/ana}

We find that pruning the second shortcut of OR-Spiking ResNet with SynA-T and the three shortcuts of OR-Spiking ResNet with SynA-C did not result in any reduction in the model's classification accuracy. This discovery stems from our analysis of the firing rates of LIF neurons in the OR-Spiking ResNet model with SynA-C attention, as depicted in Figure \ref{fig7}. Starting from a certain round of training, the firing rates of LIF neurons in the three shortcuts were reduced to 0 and remained constant, indicating that none of these three shortcuts generated any spikes. This characteristic enables pruning to decrease the model's resource consumption without compromising its performance. This phenomenon occurs naturally and does not require any manual pruning, hence we refer to it as "natural pruning". This feature opens up new possibilities for deploying the model on neuromorphic hardware and applying SNNs in edge computing.

Because VGG can outperform ResNet on some smaller datasets or datasets with easily extractable features due to the tendency of the ResNet structure to overfit, the effectiveness of our SynA in improving model classification accuracy naturally raises a question: is the improvement in classification accuracy due to ResNet degrading to VGG and not network attention learning more important features? Therefore, we conducted a set of ablation experiments to demonstrate that the performance enhancement by SynA is not due to natural pruning but rather because SynA makes the model focus on more critical features.

As shown in Table \ref{B-3-table-1}, we conducted experiments where we removed the second shortcut of OR-Spiking ResNet before training and added SynA-T at other positions as usual. The resulting model achieved only 95.83\% classification accuracy on the Dvs Gesture dataset, which is 1.74\% lower than the model obtained by natural pruning with SynA-T. Similarly, we designed the same experiment for SynA-C, and the manually pruned OR-Spiking ResNet before training had a 1.39\% lower classification accuracy compared to the model pruned with SynA-C through natural pruning. These two sets of experiments strongly demonstrate that the improvement in model classification accuracy by SynA is not due to natural pruning but rather the result of SynA's inherent attention mechanism.

\subsection{Experimental settings}
\label{sec:experimental}

\begin{table}[]

\caption{Hyper-parameters for experiments}
\label{C-table-2}
\centering
\setlength{\tabcolsep}{3pt}
\begin{tabular}{@{}ccc@{}}
\toprule
Datasets                                                                        & Name                               & Value                                                                                                          \\ \midrule
\multirow{5}{*}{DVS Gesture}                                                    & lr                                 & 1e-4                                                                                                         \\
                                                                                & T                                  & 32                                                                                                             \\
                                                                                & batch\_size                        & 32                                                                                                             \\
                                                                                & train\_epoch                       & 1000                                                                                                           \\
                                                                                & data\_transformer                  & -                                                                                                              \\ \midrule
\multirow{7}{*}{CIFAR10-DVS}                                                    & lr                                 & 1e-3                                                                                                          \\
                                                                                & T                                  & 16                                                                                                             \\
                                                                                & batch\_size                        & 128                                                                                                             \\
                                                                                & train\_epoch                       & 500                                                                                                           \\
                                                                                & \multirow{2}{*}{data\_transformer} & RandomHorizontalFlip(p=0.5)                                                                                    \\
                                                                                &                                    & \begin{tabular}[c]{@{}c@{}}RandomAffine(degrees=0,\\ translate=(2.5, 5/128))\end{tabular}                      \\ \midrule
\multirow{5}{*}{MNIST}                                                          & lr                                 & 1e-2                                                                                                           \\
                                                                                & T                                  & 16                                                                                                             \\
                                                                                & batch\_size                        & 128                                                                                                            \\
                                                                                & train\_epoch                       & 100                                                                                                            \\
                                                                                & data\_transformer                  & -                                                                                                              \\ \midrule
\multirow{6}{*}{Fashion-MNIST}                                                  & lr                                 & 1e-2                                                                                                           \\
                                                                                & T                                  & 16                                                                                                             \\
                                                                                & batch\_size                        & 128                                                                                                            \\
                                                                                & train\_epoch                       & 100                                                                                                            \\
                                                                                & \multirow{2}{*}{data\_transformer} & RandomHorizontalFlip(p=0.5)
                                                                                \\ 
                                                                                &                                    & Normalize(mean=[0.5], std=[0.5])
                                                                                \\  \bottomrule
\end{tabular}
\end{table}

All of our experiments were conducted on 4 * NVIDIA 3090 GPUs. To ensure experiment reproducibility, we have not only made our entire codebase open to the public but have also provided detailed training parameters and network microstructures for each dataset. The hyperparameter settings for each dataset are presented in Table \ref{C-table-2}, while the network parameter settings for SynA-ResNet on each dataset are outlined in Table \ref{C-table-1}. Moreover, cxkysz and MPkysz are the Conv2D and MaxPooling layer with output channels = x, kernel size = y, and stride = z. BN denotes the BatchNorm layer. LIF is the LIF neuron layer. FC denotes the fully connected layer. AP is the global average pooling layer. the \{\}*n indicates the structure repeating n times.

\section{Conclusion}
\label{sec:con}
The proposed SynA-ResNet overcomes the limitations of SEW-ResNet in performing spike-driven computation while maintaining high performance. SynA-ResNet first accumulates a large amount of redundant information through the ORRC. Subsequently, it enhances feature extraction capabilities in the backbone network and suppresses the capture of noise and irrelevant features in the shortcut connections through the SynA mechanism. To the best of our knowledge, this is the first attention mechanism in SNNs that exploits synergistic effects. Experimental results demonstrate that SynA effectively compensates for the accuracy loss caused by high quantization in ORRC. On benchmark datasets, including Dvs128 Gesture (97.92\%), CIFAR10-DVS (78.51\%), MNIST (99.63\%), and Fashion-MNIST (94.43\%), our method achieves performance comparable to or surpassing state-of-the-art (SOTA) residual SNN methods. Remarkably, our approach reduces energy consumption by up to a factor of 28 and spike count by up to a factor of 1.75, showcasing its exceptional efficiency.

In summary, this work demonstrates that high quantization may not have a catastrophic impact on model performance. Furthermore, we introduce the novel concept of inhibitory attention and utilize spiking neurons for computing attention weights, which adds an extra layer of biological plausibility to the attention mechanism. We believe that our proposed innovative training paradigm will provide valuable guidance and inspire the development of new training methods and model architectures. Moreover, our work is anticipated to stimulate research on lightweight SNN models and facilitate their deployment on neuromorphic hardware platforms, such as Intel's Loihi and IBM's TrueNorth, paving the way for more efficient and biologically-inspired artificial intelligence systems.

\printbibliography

@inproceedings{1,
  title={Deep residual learning for image recognition},
  author={He, Kaiming and Zhang, Xiangyu and Ren, Shaoqing and Sun, Jian},
  booktitle={Proceedings of the IEEE conference on computer vision and pattern recognition},
  pages={770--778},
  year={2016}
}

@inproceedings{2,
  title={Non-parametric online learning from human feedback for neural machine translation},
  author={Wang, Dongqi and Wei, Haoran and Zhang, Zhirui and Huang, Shujian and Xie, Jun and Chen, Jiajun},
  booktitle={Proceedings of the AAAI Conference on Artificial Intelligence},
  volume={36},
  number={10},
  pages={11431--11439},
  year={2022}
}

@article{4,
  title={Pali: A jointly-scaled multilingual language-image model},
  author={Chen, Xi and Wang, Xiao and Changpinyo, Soravit and Piergiovanni, AJ and Padlewski, Piotr and Salz, Daniel and Goodman, Sebastian and Grycner, Adam and Mustafa, Basil and Beyer, Lucas and others},
  journal={arXiv preprint arXiv:2209.06794},
  year={2022}
}

@article{5,
  title={Towards spike-based machine intelligence with neuromorphic computing},
  author={Roy, Kaushik and Jaiswal, Akhilesh and Panda, Priyadarshini},
  journal={Nature},
  volume={575},
  number={7784},
  pages={607--617},
  year={2019},
  publisher={Nature Publishing Group UK London}
}

@book{6,
  title={Neuronal dynamics: From single neurons to networks and models of cognition},
  author={Gerstner, Wulfram and Kistler, Werner M and Naud, Richard and Paninski, Liam},
  year={2014},
  publisher={Cambridge University Press}
}

@article{7,
  title={Networks of spiking neurons: the third generation of neural network models},
  author={Maass, Wolfgang},
  journal={Neural networks},
  volume={10},
  number={9},
  pages={1659--1671},
  year={1997},
  publisher={Elsevier}
}

@article{8,
  title={Deep residual learning in spiking neural networks},
  author={Fang, Wei and Yu, Zhaofei and Chen, Yanqi and Huang, Tiejun and Masquelier, Timoth{\'e}e and Tian, Yonghong},
  journal={Advances in Neural Information Processing Systems},
  volume={34},
  pages={21056--21069},
  year={2021}
}

@inproceedings{9,
  title={Going deeper with directly-trained larger spiking neural networks},
  author={Zheng, Hanle and Wu, Yujie and Deng, Lei and Hu, Yifan and Li, Guoqi},
  booktitle={Proceedings of the AAAI conference on artificial intelligence},
  volume={35},
  number={12},
  pages={11062--11070},
  year={2021}
}

@article{10,
  title={Advancing Spiking Neural Networks towards Deep Residual Learning},
  author={Hu, Yifan and Deng, Lei and Wu, Yujie and Yao, Man and Li, Guoqi},
  journal={arXiv preprint arXiv:2112.08954},
  year={2021}
}

@article{11,
  title={Attention spiking neural networks},
  author={Yao, Man and Zhao, Guangshe and Zhang, Hengyu and Hu, Yifan and Deng, Lei and Tian, Yonghong and Xu, Bo and Li, Guoqi},
  journal={IEEE transactions on pattern analysis and machine intelligence},
  year={2023},
  publisher={IEEE}
}

@inproceedings{12,
  title={Spiking-yolo: spiking neural network for energy-efficient object detection},
  author={Kim, Seijoon and Park, Seongsik and Na, Byunggook and Yoon, Sungroh},
  booktitle={Proceedings of the AAAI conference on artificial intelligence},
  volume={34},
  number={07},
  pages={11270--11277},
  year={2020}
}

@article{14,
  title={A New Pre-conditioned STDP Rule and Its Hardware Implementation in Neuromorphic Crossbar Array},
  author={Tao, Tuomin and Li, Da and Ma, Hanzhi and Li, Yan and Tan, Shurun and Liu, En-xiao and Schutt-Aine, Jose and Li, Er-Ping},
  journal={Neurocomputing},
  pages={126682},
  year={2023},
  publisher={Elsevier}
}

@inproceedings{15,
  title={SwiftSpike: An efficient software framework for the development of spiking neural networks},
  author={Fahey, Genevieve Claire and Ippolito, Samuel J and Matthews, Glenn I},
  booktitle={2023 IEEE International Conference on Omni-layer Intelligent Systems (COINS)},
  pages={1--6},
  year={2023},
  organization={IEEE}
}

@inproceedings{16,
  title={Fast and efficient information transmission with burst spikes in deep spiking neural networks},
  author={Park, Seongsik and Kim, Seijoon and Choe, Hyeokjun and Yoon, Sungroh},
  booktitle={Proceedings of the 56th Annual Design Automation Conference 2019},
  pages={1--6},
  year={2019}
}

@article{18,
  title={Surrogate gradient learning in spiking neural networks: Bringing the power of gradient-based optimization to spiking neural networks},
  author={Neftci, Emre O and Mostafa, Hesham and Zenke, Friedemann},
  journal={IEEE Signal Processing Magazine},
  volume={36},
  number={6},
  pages={51--63},
  year={2019},
  publisher={IEEE}
}

@article{20,
  title={Spatio-temporal backpropagation for training high-performance spiking neural networks},
  author={Wu, Yujie and Deng, Lei and Li, Guoqi and Zhu, Jun and Shi, Luping},
  journal={Frontiers in neuroscience},
  volume={12},
  pages={331},
  year={2018},
  publisher={Frontiers Media SA}
}

@article{24,
  title={Binaryconnect: Training deep neural networks with binary weights during propagations},
  author={Courbariaux, Matthieu and Bengio, Yoshua and David, Jean-Pierre},
  journal={Advances in neural information processing systems},
  volume={28},
  year={2015}
}

@article{25,
  title={Glif: A unified gated leaky integrate-and-fire neuron for spiking neural networks},
  author={Yao, Xingting and Li, Fanrong and Mo, Zitao and Cheng, Jian},
  journal={Advances in Neural Information Processing Systems},
  volume={35},
  pages={32160--32171},
  year={2022}
}

@article{28,
  title={Attention enhances synaptic efficacy and the signal-to-noise ratio in neural circuits},
  author={Briggs, Farran and Mangun, George R and Usrey, W Martin},
  journal={Nature},
  volume={499},
  number={7459},
  pages={476--480},
  year={2013},
  publisher={Nature Publishing Group UK London}
}

@inproceedings{30,
  title={Temporal-wise attention spiking neural networks for event streams classification},
  author={Yao, Man and Gao, Huanhuan and Zhao, Guangshe and Wang, Dingheng and Lin, Yihan and Yang, Zhaoxu and Li, Guoqi},
  booktitle={Proceedings of the IEEE/CVF International Conference on Computer Vision},
  pages={10221--10230},
  year={2021}
}

@article{31,
  title={Spiking deep residual networks},
  author={Hu, Yangfan and Tang, Huajin and Pan, Gang},
  journal={IEEE Transactions on Neural Networks and Learning Systems},
  year={2021},
  publisher={IEEE}
}

@article{32,
  title={Toward scalable, efficient, and accurate deep spiking neural networks with backward residual connections, stochastic softmax, and hybridization},
  author={Panda, Priyadarshini and Aketi, Sai Aparna and Roy, Kaushik},
  journal={Frontiers in Neuroscience},
  volume={14},
  pages={653},
  year={2020},
  publisher={Frontiers Media SA}
}

@inproceedings{35,
  title={Cbam: Convolutional block attention module},
  author={Woo, Sanghyun and Park, Jongchan and Lee, Joon-Young and Kweon, In So},
  booktitle={Proceedings of the European conference on computer vision (ECCV)},
  pages={3--19},
  year={2018}
}

@inproceedings{36,
  title={Temporal-wise attention spiking neural networks for event streams classification},
  author={Yao, Man and Gao, Huanhuan and Zhao, Guangshe and Wang, Dingheng and Lin, Yihan and Yang, Zhaoxu and Li, Guoqi},
  booktitle={Proceedings of the IEEE/CVF International Conference on Computer Vision},
  pages={10221--10230},
  year={2021}
}

@article{38,
  title={Training Full Spike Neural Networks via Auxiliary Accumulation Pathway},
  author={Chen, Guangyao and Peng, Peixi and Li, Guoqi and Tian, Yonghong},
  journal={arXiv preprint arXiv:2301.11929},
  year={2023}
}

@article{39,
  title={The impulses produced by sensory nerve endings: Part I},
  author={Adrian, Edgar D},
  journal={The Journal of physiology},
  volume={61},
  number={1},
  pages={49},
  year={1926},
  publisher={Wiley-Blackwell}
}

@inproceedings{40,
  title={Fast and efficient information transmission with burst spikes in deep spiking neural networks},
  author={Park, Seongsik and Kim, Seijoon and Choe, Hyeokjun and Yoon, Sungroh},
  booktitle={Proceedings of the 56th Annual Design Automation Conference 2019},
  pages={1--6},
  year={2019}
}

@misc{45,
	title = {SpikingJelly},
	author = {Fang, Wei and Chen, Yanqi and Ding, Jianhao and Yu, Zhaofei and Masquelier, Timothée and Chen, Ding and Huang, Liwei and Zhou, Huihui and Li, Guoqi and Tian, Yonghong and others},
	year = {2020},
	howpublished = {\url{https://github.com/fangwei123456/spikingjelly}},
	note = {Accessed: YYYY-MM-DD},
}

@inproceedings{46,
  title={A low power, fully event-based gesture recognition system},
  author={Amir, Arnon and Taba, Brian and Berg, David and Melano, Timothy and McKinstry, Jeffrey and Di Nolfo, Carmelo and Nayak, Tapan and Andreopoulos, Alexander and Garreau, Guillaume and Mendoza, Marcela and others},
  booktitle={Proceedings of the IEEE conference on computer vision and pattern recognition},
  pages={7243--7252},
  year={2017}
}

@article{47,
  title={Cifar10-dvs: an event-stream dataset for object classification},
  author={Li, Hongmin and Liu, Hanchao and Ji, Xiangyang and Li, Guoqi and Shi, Luping},
  journal={Frontiers in neuroscience},
  volume={11},
  pages={309},
  year={2017},
  publisher={Frontiers Media SA}
}

@article{50,
  title={Gradient-based learning applied to document recognition},
  author={LeCun, Yann and Bottou, L{\'e}on and Bengio, Yoshua and Haffner, Patrick},
  journal={Proceedings of the IEEE},
  volume={86},
  number={11},
  pages={2278--2324},
  year={1998},
  publisher={Ieee}
}

@article{51,
  title={Fashion-mnist: a novel image dataset for benchmarking machine learning algorithms},
  author={Xiao, Han and Rasul, Kashif and Vollgraf, Roland},
  journal={arXiv preprint arXiv:1708.07747},
  year={2017}
}

@article{speech_recognition,
  title={Deep spiking neural networks for large vocabulary automatic speech recognition},
  author={Wu, Jibin and Y{\i}lmaz, Emre and Zhang, Malu and Li, Haizhou and Tan, Kay Chen},
  journal={Frontiers in neuroscience},
  volume={14},
  pages={199},
  year={2020},
  publisher={Frontiers Media SA}
}

@article{53,
  title={Slayer: Spike layer error reassignment in time},
  author={Shrestha, Sumit B and Orchard, Garrick},
  journal={Advances in neural information processing systems},
  volume={31},
  year={2018}
}

@article{54,
  title={Efficient processing of spatio-temporal data streams with spiking neural networks},
  author={Kugele, Alexander and Pfeil, Thomas and Pfeiffer, Michael and Chicca, Elisabetta},
  journal={Frontiers in Neuroscience},
  volume={14},
  pages={439},
  year={2020},
  publisher={Frontiers Media SA}
}

@article{55,
  title={Liaf-net: Leaky integrate and analog fire network for lightweight and efficient spatiotemporal information processing},
  author={Wu, Zhenzhi and Zhang, Hehui and Lin, Yihan and Li, Guoqi and Wang, Meng and Tang, Ye},
  journal={IEEE Transactions on Neural Networks and Learning Systems},
  volume={33},
  number={11},
  pages={6249--6262},
  year={2021},
  publisher={IEEE}
}

@inproceedings{56,
  title={Incorporating learnable membrane time constant to enhance learning of spiking neural networks},
  author={Fang, Wei and Yu, Zhaofei and Chen, Yanqi and Masquelier, Timoth{\'e}e and Huang, Tiejun and Tian, Yonghong},
  booktitle={Proceedings of the IEEE/CVF international conference on computer vision},
  pages={2661--2671},
  year={2021}
}

@article{57,
  title={Convolutional spiking neural networks for spatio-temporal feature extraction},
  author={Samadzadeh, Ali and Far, Fatemeh Sadat Tabatabaei and Javadi, Ali and Nickabadi, Ahmad and Chehreghani, Morteza Haghir},
  journal={Neural Processing Letters},
  pages={1--17},
  year={2023},
  publisher={Springer}
}

@article{58,
  title={Dart: distribution aware retinal transform for event-based cameras},
  author={Ramesh, Bharath and Yang, Hong and Orchard, Garrick and Le Thi, Ngoc Anh and Zhang, Shihao and Xiang, Cheng},
  journal={IEEE transactions on pattern analysis and machine intelligence},
  volume={42},
  number={11},
  pages={2767--2780},
  year={2019},
  publisher={IEEE}
}

@article{59,
  title={Rethinking the performance comparison between SNNS and ANNS},
  author={Deng, Lei and Wu, Yujie and Hu, Xing and Liang, Ling and Ding, Yufei and Li, Guoqi and Zhao, Guangshe and Li, Peng and Xie, Yuan},
  journal={Neural networks},
  volume={121},
  pages={294--307},
  year={2020},
  publisher={Elsevier}
}

@article{lif,
  title={A logical calculus of the ideas immanent in nervous activity},
  author={McCulloch, Warren S and Pitts, Walter},
  journal={The bulletin of mathematical biophysics},
  volume={5},
  pages={115--133},
  year={1943},
  publisher={Springer}
}

@article{spikingformer_1,
  title={Spikingformer: Spike-driven Residual Learning for Transformer-based Spiking Neural Network},
  author={Zhou, Chenlin and Yu, Liutao and Zhou, Zhaokun and Zhang, Han and Ma, Zhengyu and Zhou, Huihui and Tian, Yonghong},
  journal={arXiv preprint arXiv:2304.11954},
  year={2023}
}

@article{shan2024advancing,
  title={Advancing Spiking Neural Networks towards Multiscale Spatiotemporal Interaction Learning},
  author={Shan, Yimeng and Zhang, Malu and Zhu, Rui-jie and Qiu, Xuerui and Eshraghian, Jason K and Qu, Haicheng},
  journal={arXiv preprint arXiv:2405.13672},
  year={2024}
}

@article{zhu2022tcja,
  title={TCJA-SNN: Temporal-channel joint attention for spiking neural networks. arXiv},
  author={Zhu, RJ and Zhao, Q and Zhang, T and Deng, H and Duan, Y and Zhang, M and Deng, LJ},
  journal={arXiv preprint arXiv:2206.10177},
  year={2022}
}

@inproceedings{kim2022rate,
  title={Rate coding or direct coding: Which one is better for accurate, robust, and energy-efficient spiking neural networks?},
  author={Kim, Youngeun and Park, Hyoungseob and Moitra, Abhishek and Bhattacharjee, Abhiroop and Venkatesha, Yeshwanth and Panda, Priyadarshini},
  booktitle={ICASSP 2022-2022 IEEE International Conference on Acoustics, Speech and Signal Processing (ICASSP)},
  pages={71--75},
  year={2022},
  organization={IEEE}
}

@inproceedings{wu2019direct,
  title={Direct training for spiking neural networks: Faster, larger, better},
  author={Wu, Yujie and Deng, Lei and Li, Guoqi and Zhu, Jun and Xie, Yuan and Shi, Luping},
  booktitle={Proceedings of the AAAI conference on artificial intelligence},
  volume={33},
  number={01},
  pages={1311--1318},
  year={2019}
}

@article{davies2018loihi,
  title={Loihi: A neuromorphic manycore processor with on-chip learning},
  author={Davies, Mike and Srinivasa, Narayan and Lin, Tsung-Han and Chinya, Gautham and Cao, Yongqiang and Choday, Sri Harsha and Dimou, Georgios and Joshi, Prasad and Imam, Nabil and Jain, Shweta and others},
  journal={Ieee Micro},
  volume={38},
  number={1},
  pages={82--99},
  year={2018},
  publisher={IEEE}
}

@article{akopyan2015truenorth,
  title={Truenorth: Design and tool flow of a 65 mw 1 million neuron programmable neurosynaptic chip},
  author={Akopyan, Filipp and Sawada, Jun and Cassidy, Andrew and Alvarez-Icaza, Rodrigo and Arthur, John and Merolla, Paul and Imam, Nabil and Nakamura, Yutaka and Datta, Pallab and Nam, Gi-Joon and others},
  journal={IEEE transactions on computer-aided design of integrated circuits and systems},
  volume={34},
  number={10},
  pages={1537--1557},
  year={2015},
  publisher={IEEE}
}

@article{zhu2023spikegpt,
  title={Spikegpt: Generative pre-trained language model with spiking neural networks},
  author={Zhu, Rui-Jie and Zhao, Qihang and Li, Guoqi and Eshraghian, Jason K},
  journal={arXiv preprint arXiv:2302.13939},
  year={2023}
}

@article{zhou2022spikformer,
  title={Spikformer: When spiking neural network meets transformer},
  author={Zhou, Zhaokun and Zhu, Yuesheng and He, Chao and Wang, Yaowei and Yan, Shuicheng and Tian, Yonghong and Yuan, Li},
  journal={arXiv preprint arXiv:2209.15425},
  year={2022}
}

\end{document}